\documentclass[letterpaper, 10 pt, conference]{ieeeconf}  

\IEEEoverridecommandlockouts                              

\usepackage{graphics} 
\usepackage{times} 
\usepackage{amssymb}  
\usepackage{siunitx}
\DeclareSIUnit\fps{fps}
\usepackage{amsmath,amsfonts}
\usepackage{array}
\usepackage{textcomp}
\usepackage{stfloats}
\usepackage{url}
\usepackage{verbatim}
\usepackage{graphicx}
\usepackage{booktabs}
\usepackage{bm}
\usepackage{cite}

\usepackage{threeparttable} 
\usepackage{multirow}
\usepackage{color}
\usepackage{float}

\usepackage[dvipsnames]{xcolor}

\usepackage{balance}

\title{
\LARGE \bf
2D GauSS-MI: Efficient Active Scene Reconstruction with 
\\
Balanced Visual and Geometric Quality
}

\author{
Yuhan Xie and Jia Pan
\thanks{
The authors are with the School of Computing and Data Science, the University of Hong Kong, Hong Kong SAR, China (Email: yuhanxie@connect.hku.hk, jpan@cs.hku.hk). 
}
}

\begin{document}

\maketitle
\thispagestyle{empty}
\pagestyle{empty}

\begin{abstract}

Active reconstruction requires efficient active view selection to achieve high-quality reconstruction within limited onboard computational resources. Existing methods face challenges in adequately balancing visual and geometric quality with the computational efficiency required for real-time operation. In this work, we present an active reconstruction framework based on 2D Gaussian Splatting (2DGS). We develop an efficient online 2DGS mapping pipeline for incremental RGB-D observations and introduce a probabilistic reliability model that characterizes the view-dependent reconstruction quality of individual 2D Gaussian splats. Building on this model, we formulate 2D Gaussian Splatting Shannon Mutual Information (2D GauSS-MI), a mutual-information-based metric that exploits the explicit surface orientation of 2DGS to evaluate the expected information gain of candidate views. The proposed metric enables active view selection to account for both visual and geometric reconstruction quality. We evaluate the proposed system against three state-of-the-art baselines on eight Replica scenes. Experimental results demonstrate that our method achieves a favorable balance between visual and geometric reconstruction quality with substantially lower computational cost and competitive model storage.


\end{abstract}


\section{Introduction}

Radiance-field-based 3D reconstruction has attracted significant attention in recent years due to its high visual fidelity and fast rendering capability~\cite{mildenhall2021nerf, kerbl20233d}. 
Recent advances in radiance-field representations have further improved geometric reconstruction accuracy while preserving their strong rendering performance, exemplified by 
Gaussian Surfels~\cite{Dai2024GaussianSurfels} and 2D Gaussian Splatting (2DGS)~\cite{Huang2DGS2024}.
These advances have also motivated the development of active reconstruction and exploration for robotic systems, which enable robots to autonomously acquire informative observations and progressively improve their scene understanding.
Such active perception capabilities can substantially enhance robotic autonomy in applications such as search and rescue, construction, and inspection. 
However, compared with conventional 3D reconstruction, active reconstruction introduces additional system-level requirements, including online reconstruction, next-best-view (NBV) selection, robot planning, and computational efficiency for resource-limited onboard platforms.

Prior to the development of efficient radiance-field reconstruction methods, active reconstruction and exploration approaches predominantly relied on computationally efficient geometric representations, such as voxel grids, meshes, and point clouds. 
Consequently, their NBV selection criteria were primarily designed to assess geometric information, such as surface coverage or geometric uncertainty~\cite{zhang2020fsmi}.
Early radiance-field-based active reconstruction methods naturally inherited these geometry-driven NBV criteria~\cite{yan2023active}. 
With the adoption of radiance-field representations that enable efficient novel-view rendering, recent works have begun to exploit their rendering capability to directly evaluate visual quality for NBV selection~\cite{YuhanRSS25}. 
Nevertheless, existing methods still face challenges in balancing visual fidelity with geometric quality, modeling view-dependent information gain for NBV evaluation, and maintaining computational efficiency for online view selection.

\begin{figure}[t] 
    \centering
    \includegraphics[width=0.97\linewidth]{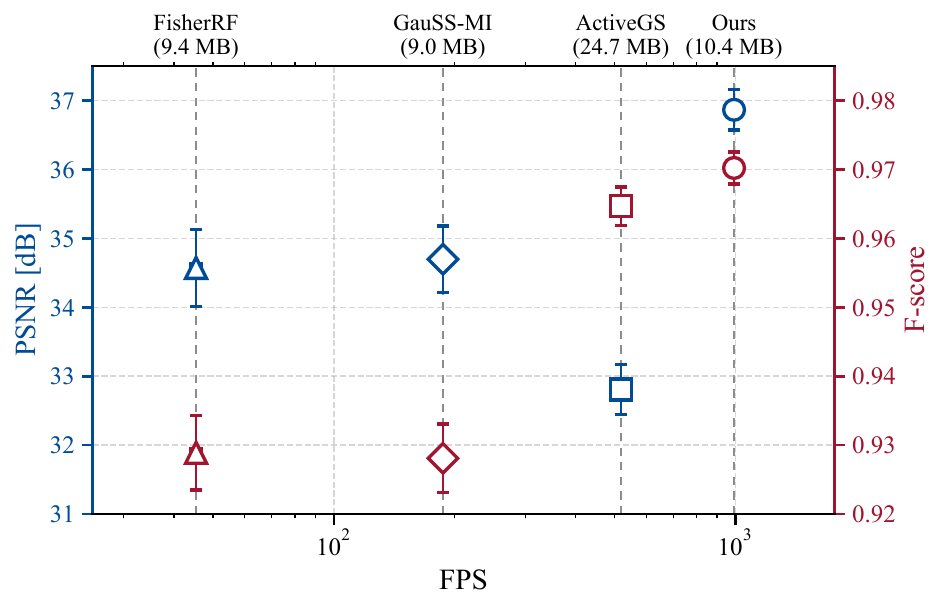}
    \vspace{-4mm}
    \caption{
        Overall comparison of reconstruction quality, computational efficiency, and model size. The numbers in parentheses indicate the model size.
        Results are averaged over three runs for each of the eight scenes.
        Our method achieves a favorable balance among visual and geometric quality, computational efficiency, and model compactness.
    }
    \label{fig_sactter_methods}
    \vspace{-4mm}
\end{figure}

To overcome the aforementioned challenges, this paper proposes an efficient active reconstruction system based on 2DGS~\cite{Huang2DGS2024}, which jointly considers visual and geometric quality throughout online reconstruction and NBV selection. 
Specifically, we develop an efficient online 2DGS reconstruction module that optimizes rendering quality with geometry-aware regularization.
We further introduce a view-dependent probabilistic reliability model to characterize the observation quality of individual 2D Gaussian splats, and formulate an efficient NBV metric based on the resulting reliability. 
Finally, we integrate these components into a complete active reconstruction system that achieves high reconstruction quality with computational efficiency and a compact model.
The implementation of the proposed system will be publicly released\footnote{GitHub link: *} upon acceptance to facilitate future research in active reconstruction.

The main contributions of our work are summarized as follows:
\begin{itemize}
    \item A probabilistic reliability model for 2DGS that quantifies the uncertainty of the reconstructed map.
    \item A 2D Gaussian Splatting Shannon Mutual Information (2D GauSS-MI) metric for real-time novel-view assessment, jointly considering visual and geometric reconstruction quality.
    \item An efficient active scene reconstruction system based on the probabilistic reliability model and 2D GauSS-MI.
    \item Extensive experiments across eight scenes against state-of-the-art methods, demonstrating superior visual fidelity and geometric accuracy, along with high computational efficiency and competitive model storage.
\end{itemize}

\section{Related Work}

In contrast to conventional reconstruction, where the observation sequence is given in advance, active reconstruction aims to build a complete and high-quality 3D representation by actively controlling the sensor trajectory to collect informative observations. 
The core problem is therefore to determine the next-best-view (NBV) that provides the most useful information for reconstruction. 
Early approaches primarily relied on explicit geometric representations, evaluating candidate views according to geometric information gain~\cite{isler2016information,lu2024semantics}, mutual information~\cite{zhang2020fsmi,julian2014mutual}, frontier, or surface coverage~\cite{zhang2024falcon}. 
These methods are effective for geometric exploration, but do not directly account for the visual fidelity of the reconstructed scene.

The emergence of Neural Radiance Fields (NeRF)~\cite{mildenhall2021nerf} and 3D Gaussian Splatting (3DGS)~\cite{kerbl20233d} has substantially improved the visual quality and rendering efficiency of scene reconstruction, motivating their adoption in active reconstruction. 
Early radiance-field-based active reconstruction methods still selected views primarily according to geometric completeness or coverage~\cite{yan2023active, li2025activesplat}. %
To exploit the learned representation, subsequent methods estimated view-dependent uncertainty using neural networks~\cite{pan2022activenerf,ran2023neurar}, but their performance can depend on training data and the quality of the learned uncertainty estimator. 
In contrast to NeRF's implicit modeling approach, 3DGS explicitly represents the scene with spatially distributed Gaussians, enabling uncertainty to be evaluated more directly~\cite{jiang2025fisherrf, shen2025auto3r, jun2026sa}. 

Despite their strong rendering capability, the surface reconstruction using 3D Gaussian modeling and splatting faces several challenges~\cite{Huang2DGS2024}.
The volumetric radiance representation of 3D Gaussians is not well suited to the thin nature of surfaces, and it does not natively provide surface normals, which are essential for high-quality surface reconstruction. 
Moreover, the rasterization of 3D ellipsoids lacks multi-view consistency, resulting in varied 2D intersection planes from different viewpoints.
This motivates 2D Gaussian representations, including Gaussian Surfels~\cite{Dai2024GaussianSurfels} and 2DGS~\cite{Huang2DGS2024}, which were proposed independently. 
Gaussian Surfels flatten a 3D Gaussian by constraining its $z$-scale to zero, whereas 2DGS represents each primitive as an oriented planar Gaussian with perspective-correct ray-splat intersection and geometry-aware regularization.
Both representations improve the geometric accuracy of Gaussian-based reconstruction.
Recent works have also explored Gaussian-surfel representations for active reconstruction. 
ActiveGS~\cite{jin2025ral} combines Gaussian surfels with a voxel map and uses Gaussian confidence to identify under-reconstructed regions. 
ObjSplat~\cite{li2026objsplat} introduces geometry-aware view evaluation for active object reconstruction based on back-face visibility and multi-view covisibility. 

Recent Gaussian-Surfel-based active reconstruction methods demonstrate the potential of 2D Gaussian representations for active perception. 
However, view-dependent quality assessment requires not only a surface-oriented representation but also geometrically consistent rendering across viewpoints, which is particularly important for NBV selection where candidate views are unobserved and often oblique.
Both 3DGS and Gaussian Surfels employ an affine approximation of perspective projection, whose accuracy degrades away from the Gaussian center. 
In contrast, 2DGS achieves multi-view-consistent surface geometry through perspective-correct ray-splat intersection~\cite{Huang2DGS2024}. 
Moreover, unlike Gaussian surfels, which require external normal priors during training and volumetric post-processing, 2DGS only incorporates geometry-aware regularization during training, making it more suitable for RGB-D-based active reconstruction.

Building on these properties, we adopt the information-theoretic framework of GauSS-MI~\cite{YuhanRSS25} and reformulate it based on 2DGS.
Specifically, GauSS-MI defines view-dependent reliability over four horizontal quadrants in the world frame, which does not explicitly account for the orientation of individual Gaussians. 
The explicit surface normal of 2DGS instead provides a natural reference for defining reliability with respect to each primitive's intrinsic orientation. 
We therefore define an orientation-aware reliability model and formulate the 2D GauSS-MI metric accordingly, 
enabling view selection to better capture the information contributed by different observations. 
Furthermore, we leverage the geometry-aware properties of 2DGS to jointly incorporate visual and geometric quality into NBV selection and online mapping optimization, leading to improved reconstruction results.
To the best of our knowledge, this is the first active scene reconstruction system built upon the 2DGS representation.


\begin{table}[t]

\centering
\caption{Main Notations for Methodology}
\label{tab_notation}
\setlength{\tabcolsep}{7pt}
\begin{threeparttable}

\begin{tabular}{@{}ll@{}}
    \toprule
Notations & Explanation \\ 
    \midrule
    $\mathcal{G}$         & 2D Gaussian Splatting map. \\
    $\bm\sigma$         & Camera pose, or viewpoint.       \\
    $\bm\mu$             & Position of a Gaussian. \\
    $\bm n$             & Normal vector of a Gaussian. \\
    $T$             & Cumulative transmittance of a Gaussian. \\
    ${C}, \hat{C}$             & Rendered color and observed color. \\
    ${D}, \hat{D}$             & Rendered depth and observed depth. \\
    ${\bm N}, \hat{\bm N}$             & Rendered normal and normal estimated from observed depth. \\
    ${\bm N}_{{D}}$             & Normal estimated from rendered depth. \\

    $\lambda$             & Hyperparameters. \\
    $\mathcal{L}$             & Training loss. \\
    $L$             & Reconstruction loss for reliability update. \\
    $P(r)$             & Reliability probability of a Gaussian. \\
    $o, l$             & Odds and log odds probability of a Gaussian. \\
    $\delta$             & Inverse sensing model. \\
    $z, Z$				& Random variable and realization of an observation. \\
    $I$             & Mutual information. \\ 
    ${(\cdot)}^{[i]}$           & Index of a Gaussian. \\
    ${(\cdot)}^{[j]}$           & Index of a pixel. \\
    ${(\cdot)}^{[m]}$           & Index of an ordered Gaussian. \\
    ${(\cdot)}_k$           & Property based on the observation at time $k$. \\
    ${(\cdot)}_{1:k}$           & Property based on observations from the start to time $k$. \\
    ${(\cdot)}_n$           & Property of candidate viewpoint $n$. \\
    \bottomrule
\end{tabular}

\end{threeparttable}

\vspace{-0.4cm}
\end{table}

\begin{figure}[t] 
    \centering
    \includegraphics[width=\linewidth]{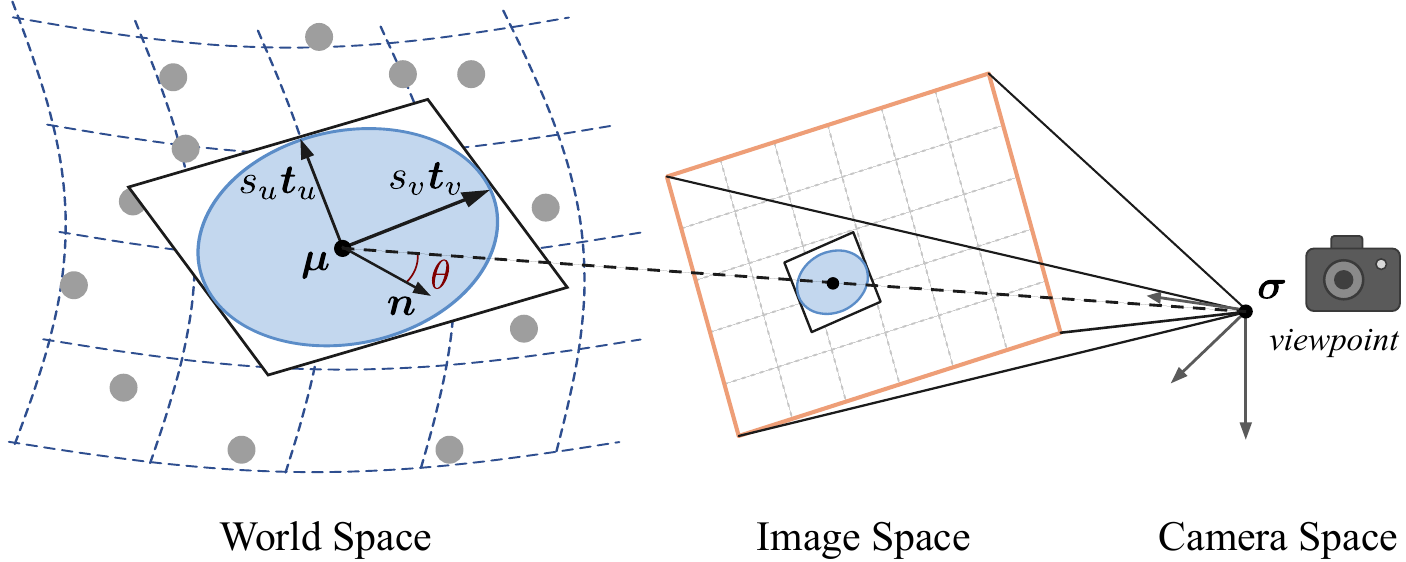}
    \vspace{-4mm}
    \caption{
        Illustration of 2D Gaussian splatting.
    }
    \label{fig_2dgs_proj}
    \vspace{-6mm}
\end{figure}

\section{Mapping}
\label{sec_mapping}

This section introduces the map representation of our active reconstruction system, 2DGS~\cite{Huang2DGS2024}, and the online reconstruction pipeline that processes incremental observations in real time. 
The resulting online map provides the basis for active view evaluation and selection in Section~\ref{sec_active_view}.
The main notations used throughout the methodology are summarized in Table~\ref{tab_notation}.

\subsection{2D Gaussian Splatting}
\label{subsec_2dgs}

As illustrated in Figure~\ref{fig_2dgs_proj}, a 2D Gaussian is defined on a local tangent plane by its center $\bm\mu$ and two orthogonal tangential vectors $\bm t_u, \bm t_v$ with scale factors $s_u, s_v$.
The unit normal vector is given by $\bm n = \bm t_u \times \bm t_v$, forming the rotation matrix $\bm R = [\bm t_u, \bm t_v, \bm n]$.
Each Gaussian also carries an opacity $\alpha$ and a color $c$~\cite{Huang2DGS2024}. 
To reduce computational overhead, spherical harmonics are omitted, making the color view-independent.

To render an image from a viewpoint $\bm\sigma$, 
the Gaussians intersecting each camera ray are ordered by depth and composited front-to-back.
The cumulative transmittance of Gaussian $i$ is defined as
\begin{equation}    \label{eq_cumu_transmit}
T^{[i]}= \alpha^{[i]} \prod_{m<i} (1-\alpha^{[m]})
\end{equation}
The rendered color $C$, depth $D$, and normal $\bm N$ images are then computed as
\begin{equation}    \label{eq_raster}
    C = \sum_i c^{[i]}\, T^{[i]}, \ 
    D = \sum_i d^{[i]}\, T^{[i]}, \ 
    \bm N = \sum_i \bm n^{[i]}\, T^{[i]}
\end{equation}
where $d^{[i]}$ denotes the depth of the ray-splat intersection.

\subsection{Incremental Online Reconstruction}

Online reconstruction is essential for active reconstruction, as it incrementally updates the Gaussian map in real time and maintains an up-to-date representation for subsequent active view evaluation and selection.

At each timestep $k$, the robot collects a color image $\hat C_k$ and a depth image $\hat D_k$ at an estimated camera pose $\bm\sigma_k$.
Given the observation and pose, we update the 2DGS map with new Gaussians whose centers are initialized by back-projecting the observed depth, while the normal $\bm n$ is estimated from the local depth gradient and oriented toward the observing camera. 
To bound the size of the Gaussian map, a hash voxel grid is maintained to reject newly initialized primitives whose centers lie within the distance $\gamma$ of existing Gaussians~\cite{zhong2026tro}.

The updated 2DGS map is then optimized on the new observation for a fixed number of iterations $K_{\mathrm {new}}$, 
followed by randomly replaying previously collected frames for $K_{\mathrm {replay}}$ iterations to prevent degradation of previously reconstructed regions.
To ensure that the Gaussian map not only produces high-fidelity image renderings but also captures the geometric structure of the scene, we incorporate two geometric regularizers from 2DGS~\cite{Huang2DGS2024} in addition to the rendering losses.
The first is the depth distortion loss, which concentrates Gaussians along each ray within a thin band around the surface:
\begin{equation}    \label{eq_loss_depth_dist}
    \mathcal{L}_\mathrm{D_{dist}} = \sum_{i}\sum_{m<i} T^{[i]}T^{[m]} \left| d^{[i]} - d^{[m]} \right|
\end{equation}
The second is the normal consistency loss, which further aligns the splat normals with the surface geometry estimated from rendered depth:
\begin{equation}    \label{eq_norm_cons}
    \mathcal{L}_\mathrm{N_{const}} = 1-{\bm N}^T{\bm N}_D
\end{equation}
where ${\bm N}_D$ denotes the normal estimated from the rendered depth $D$.
The training loss is thus formulated as the weighted sum of the rendering losses and geometric regularizers:
\begin{equation}    \label{eq_train_loss}
    \mathcal{L} = 
    \mathcal{L}_\mathrm{C} + 
    \lambda_\mathrm{t,D}\mathcal{L}_\mathrm{D} + 
    \lambda_\mathrm{t,D_d}\mathcal{L}_\mathrm{D_{dist}} + 
    \lambda_\mathrm{t,N_c}\mathcal{L}_\mathrm{N_{const}}
\end{equation}
where $\mathcal{L}_\mathrm{C}$ is the color rendering loss combining the $\mathcal{L}_1$ and D-SSIM terms, and $\mathcal{L}_\mathrm{D}$ is the depth rendering loss computed using the $\mathcal{L}_1$ distance.



\section{Active View Selection}
\label{sec_active_view}

Given the real-time map reconstructed in the previous section, we first develop a probabilistic reliability model for 2D Gaussian Splatting to quantify the observed information in Section~\ref{subsec_reliability}, and then formulate 2D Gaussian Splatting Shannon Mutual Information (2D GauSS-MI) for active view selection in Section~\ref{subsec_GSSMI}.

\subsection{2D Gaussian Splat Reliability}
\label{subsec_reliability}

\subsubsection{Bayesian Modeling and Update}
\label{subsubsec_bayes}
To model the information that a random observation $z$ provides to the 2DGS map $\mathcal{G}$, a random variable $r^{[i]}$ is introduced for each Gaussian splat $i\in\mathcal{G}$.
The probability that a Gaussian $i$ is \textit{reliable} for reconstruction is defined as $P(r^{[i]})\in(0,1)$, which can be equivalently represented by its odds ratio $o^{[i]}\in(0,+\infty)$ or log odds $l^{[i]}\in(-\infty,+\infty)$:
\begin{equation}\label{eq_pol_define}
    l^{[i]} := \log(o^{[i]}) := 
    \log\left(
        \frac{  P(r^{[i]})  }{  1-P(r^{[i]}) }
    \right)
\end{equation}

Initially, no prior information about the scene is assumed, and thus $P_0(r^{[i]})=0.5,\ \forall i\in\mathcal{G}$.
The reliability probabilities are assumed to be independent across Gaussian splats.
Under this assumption, $P(r^{[i]}\mid Z_{1:k})$ is recursively updated based on observations $Z_1$ to $Z_k$ using a binary Bayesian filter:
\begin{equation}
\label{eq_prob_update} 
    o^{[i]} (Z_{1:k}) 
    = \delta^{[i]} (Z_k)    o^{[i]} (Z_{1:k-1})
\end{equation}
where $\delta^{[i]}(Z_k)$ is termed the odds ratio of the inverse sensing model, 
characterizing the reliability evidence provided by observation $k$ for Gaussian $i$.
For brevity, we use $\delta^{[i]}_k$ to denote $\delta^{[i]}(Z_k)$ in the following.
The inverse sensing model is formulated based on the reconstruction loss $L_k$~\cite{YuhanRSS25},
\begin{equation}
\label{eq_inv_sensor}
    \delta^{[i]}_k
        = (\lambda_L L_k)^{-\lambda_TT^{[i]}}
\end{equation}
where $\lambda_L$ and $\lambda_T$ are hyperparameters. 
Substituting \eqref{eq_inv_sensor} into \eqref{eq_prob_update} and taking the logarithm gives 
\begin{equation}
\label{eq_imple_update_prob} 
    l_{1:k}^{[i]} 
    = 
    l_k^{[i]} + l_{1:k-1}^{[i]}
    =
    -\lambda_T T^{[i]}\log(\lambda_L L_k)    
    +
    l_{1:k-1}^{[i]}
\end{equation}
Therefore, $P(r^{[i]}\mid Z_{1:k})$ can be updated by propagating the logarithm of the reconstruction loss image $L_k$ to each Gaussian according to its cumulative transmittance $T^{[i]}$. 

\subsubsection{Reliability for 2D Gaussian Splats}
We next detail the reliability model for 2D Gaussian Splats, including its direction-dependent formulation and computation.


The update in \eqref{eq_imple_update_prob} assigns a single reliability probability to each Gaussian.
However, a Gaussian that reconstructs well from one side can be unreliable from the other, making its reliability inherently direction-dependent.
GauSS-MI~\cite{YuhanRSS25} for 3DGS addresses this with a quadrant reliability defined in the world frame along the horizontal directions, which hardly accommodates the distinct orientation of each Gaussian.
In contrast, each 2D Gaussian has an explicit normal vector that naturally defines its facing direction.
Therefore, a two-sided reliability is defined for each 2D Gaussian, corresponding to its front and back sides:
\begin{equation}
    \bm P(r^{[i]}) = 
    \left(
        P(r^{[i]}_{\mathrm{f}}),
        P(r^{[i]}_{\mathrm{b}})
    \right)
\end{equation}
The \textit{front} side is oriented along the Gaussian normal and faces the viewpoint from which it is initially observed, while the \textit{back} side corresponds to the opposite viewing direction.
In general, only the front-side reliability is meaningful, whereas the back-side reliability also becomes meaningful for thin structures that can be correctly observed from both sides.

To determine the observed side and quantify the observation weight, 
we define the unit viewing direction from Gaussian $i$ to the camera pose $\bm\sigma_k$ as
$\bm d^{[i]}_k = \left(\bm\sigma_k-\bm\mu^{[i]}\right)/\lVert \bm\sigma_k - \bm\mu^{[i]} \rVert$.
The incidence angle $\theta^{[i]}_k$ and the \textit{observation-quality weight} $w^{[i]}_k$ are defined as
\begin{equation}
\label{eq_ob_quality_w}
    w_{k}^{[i]} 
    = 
    \cos{\theta^{[i]}_k} 
    =  
    \bm d^{[i]}_k \cdot \bm n^{[i]}
\end{equation}
Thus, $w^{[i]}_k=1$ indicates a head-on observation of the front side, $w^{[i]}_k=0$ an edge-on observation, and $w^{[i]}_k<0$ an observation from the back side.
The weight $w_{k}^{[i]}$ therefore characterizes the directional discrepancy between the viewpoint $\bm\sigma_k$ and the normal of the 2D Gaussian $i$.

Since an observation only provides reliability evidence for the side it actually observes, 
the log odds to be updated is selected according to the sign of $w^{[i]}_k$:
\begin{equation}\label{eq_obs_logodds}
l_{k, \mathrm{obs}}^{[i]} = 
    \begin{cases}
        l_{k, \mathrm{f}}^{[i]}, & w^{[i]}_k > w_{\min} \\
        l_{k, \mathrm{b}}^{[i]}, & w^{[i]}_k < -w_{\min} \\
        \mathrm{skip\;update}, & |w^{[i]}_k| \le w_{\min}
    \end{cases}
\end{equation}
where the gate $w_{\min}>0$ skips Gaussians viewed nearly edge-on.

Based on the inverse sensing model in \eqref{eq_inv_sensor} and \eqref{eq_imple_update_prob}, the selected log odds are updated using the per-pixel reconstruction loss $L^{[j]}_k$ as
\begin{equation}\label{eq_reli_update_impl}
    l_{k, \mathrm{obs}}^{[i]}
    =
    \lambda_T |w_{k}^{[i]} |
    \sum_j 
    -\log(\lambda_L L^{[j]}_k )
    T^{[i,j]}
\end{equation}
Here, $T^{[i,j]}$ denotes the cumulative transmittance of Gaussian $i$ at pixel $j$, so only pixels to which Gaussian $i$ contributes are included in the update.
The occluded or out-of-view Gaussians therefore receive no update.

In \eqref{eq_reli_update_impl}, the reconstruction loss image $L_k$ combines color, depth, and normal discrepancies. For brevity, the subscript $k$ is omitted below.
\begin{equation}
  L =\mathcal{L}_{\mathrm{C} }
  + \lambda_\mathrm{r,D} \mathcal{L}_{\mathrm{D} }
  + \lambda_\mathrm{r,N_r} L_{\mathrm{N_{rend}}}
\end{equation}
where the rendered-normal loss $L_{\mathrm{N_{rend}}}$ is defined as
\begin{equation}    \label{eq_norm_rend}
    L_{\mathrm{N_{rend}}}= 1-{\bm N}^T\hat{\bm N}
\end{equation}
Here, $\hat{\bm N}$ denotes the normal estimated from the observed depth.
Note that $L_{\mathrm{N_{rend}}}$ differs from the normal-consistency loss $\mathcal{L}_\mathrm{N_{const}}$ in \eqref{eq_norm_cons}: 
the latter aligns the rendered normal with the geometry estimated from the rendered depth during training, 
whereas the former measures the discrepancy between the rendered normal and the observed geometry for reliability estimation.
The definitions of $\mathcal{L}_{\mathrm{C}}$ and $\mathcal{L}_{\mathrm{D}}$ are the same as those in \eqref{eq_train_loss}, representing the color and depth rendering losses, respectively.



\subsection{2D Gaussian Splatting Shannon Mutual Information}
\label{subsec_GSSMI}

Based on the proposed reliability model and Shannon mutual information, we formulate 2D Gaussian Splatting Shannon Mutual Information (2D GauSS-MI) to assess the information provided by novel viewpoints for NBV selection.

\subsubsection{Expected Mutual Information for Novel Viewpoints}
Given past observations $Z_{1:k-1}$, the next best view $z_k^{*}$ is selected to maximize the mutual information (MI) $I(r; z_{k}|Z_{1:k-1})$~\cite{julian2014mutual}. 
For brevity, we write $I(r;z)$ for $I(r; z_{k}|Z_{1:k-1})$ in the following.
Assuming pixel-wise independent measurements, the MI decomposes over pixels $j$ and Gaussians $i$ as
\begin{equation}\label{eq_I_decomp}
     I(r;z)=\sum_j\sum_i I(r^{[i]};z^{[j]})\,T^{[i,j]}
\end{equation}

Based on information theory~\cite{cover1991information, zhang2020fsmi}, the MI between two random variables is defined as
\begin{equation}
\begin{aligned}    
    \label{eq_MI_definition} 
    I(r^{[i]} ; z^{[j]})   
    :=&
    P(r^{[i]}, z^{[j]}=Z) 
    \log\left(
        \frac{ P(r^{[i]}, z^{[j]}=Z )  }
             { P(r^{[i]} )   P(z^{[j]}=Z  ) } 
    \right)
\\ = &
    P(z^{[j]}=Z)f(\delta^{[i]}(Z), o^{[i]}_{1:k-1}) 
\end{aligned}
\end{equation}
Here, $P(z^{[j]}=Z)$ depends only on the observation and is affected by sensor quality, and is therefore termed the \textit{measurement prior}.
The function $f(\delta^{[i]}(Z), o^{[i]}_{1:k-1})$ is termed the \textit{information gain function}, which can be derived and abbreviated as
\begin{equation}  
    f(\delta, o) = 
    \frac{ o }{o+\delta^{-1}} 
    \log\left(
        \frac{o+1}{o+\delta^{-1}}
    \right)
\end{equation}

For a novel candidate viewpoint, the corresponding observation is unavailable, and thus the inverse sensing model must be estimated in expectation.
Assuming that the expected reconstruction is reliable, we set $L_k=0$, which gives $\delta^{-1}=0$ in $f(\delta,o)$.
The information gain function then becomes
\begin{equation}    
    \label{eq_info_gain_comp}
    f^{[i]}
    = \log\left(\frac{o^{[i]}+1}{o^{[i]}}\right)
    = -\log\left(P(r^{[i]})\right)
\end{equation}

Accordingly, the expected MI for a novel viewpoint is computed as
\begin{equation}   
\label{eq_MI_compute} 
    I(r; z) =
    \sum_j P(z^{[j]})
    \sum_i -\log\left(P(r^{[i]})\right) T^{[i,j]}
\end{equation}

\subsubsection{2D GauSS-MI}
Similar to the reliability update in \eqref{eq_reli_update_impl}, the observation-quality weight $w_n^{[i]}$ defined in \eqref{eq_ob_quality_w} is also used to characterize the directional discrepancy between the candidate viewpoint $n$ and the orientation of the 2D Gaussian $i$.
Specifically, the information gain function $f$ is computed as
\begin{equation}    \label{eq_info_gain_impl}
    f^{[i]}_{n, \mathrm{obs}}
    = -|w_n^{[i]}|\log\left(P(r_{n,\mathrm{obs}}^{[i]})\right)
\end{equation}
where $P(r_{n,\mathrm{obs}}^{[i]})$ is selected from the two-sided reliability according to \eqref{eq_obs_logodds}.

The measurement prior $P(z^{[j]})$ is approximated as $1$ in simulation, assuming that the simulator provides ground-truth observations.
For real-world deployment, $P(z^{[j]})$ can be estimated from the camera noise model following~\cite{YuhanRSS25}.

Overall, the 2D GauSS-MI for novel-view evaluation is formulated as
\begin{equation}
\label{eq_MI_compute2}
    I_n(r; z) =
    \sum_j P(z^{[j]} )
    \sum_i -|w_n^{[i]}|
    \log\left(P(r_{n,\mathrm{obs}}^{[i]})\right) T^{[i,j]}
\end{equation}
This formulation jointly accounts for the measurement noise, the current reliability of each Gaussian, and its directional discrepancy with the candidate viewpoint.


\section{Experiments}

\subsection{Experimental Setup}

The simulation environment is built on the Habitat simulator~\cite{habitat19iccv} with scenes from the Replica dataset~\cite{replica19arxiv}. 
A quadrotor serves as the agent for active reconstruction, and its trajectories are planned by SUPER~\cite{ren2025safety}, which generates dynamically feasible trajectories for the quadrotor.
The quadrotor is equipped with an RGB-D camera that captures images at a resolution of $640 \times 480$ with a $90^\circ$ field of view (FOV).
The online reconstruction module is built upon 2DGS-SLAM~\cite{zhong2026tro}, which is adapted and extended for our active reconstruction system.
We further accelerate the 2DGS rasterizer with several implementation-level optimizations to improve computational efficiency.
Both the proposed active reconstruction system and the simulator operate on a desktop with a 32-thread i9-14900K CPU and an RTX 4090 GPU. 

\subsubsection{Baselines}

The proposed method is compared against three active reconstruction baselines: ActiveGS~\cite{jin2025ral}, GauSS-MI~\cite{YuhanRSS25}, and FisherRF~\cite{jiang2025fisherrf}.
ActiveGS combines a coarse voxel map with Gaussian surfels~\cite{Dai2024GaussianSurfels} and selects viewpoints based on unexplored regions identified from the voxel map and confidence estimates from the Gaussian surfels. 
GauSS-MI constructs a probabilistic model of the 3DGS map from rendering residuals and selects the next-best view by maximizing the expected Shannon mutual information between the map and the candidate viewpoints.
FisherRF is a radiance-field-based active view selection approach that quantifies the expected information gain through the Fisher information matrix.

\subsubsection{Implementation Details}

To complete the active reconstruction system, we implement a viewpoint primitive library to generate candidate viewpoints, following~\cite{YuhanRSS25}.
At each timestep, 198 candidate viewpoints are sampled with varying 3D positions and yaw orientations. 
A trajectory to each candidate is planned by SUPER~\cite{ren2025safety} with a feasibility check prior to the NBV evaluation and selection.
GauSS-MI and FisherRF adopt the same view planner as the proposed system, and thus share the same set of candidate viewpoints for selection.
Since FisherRF provides only a view-selection criterion rather than a complete system, we integrate it into the GauSS-MI implementation, allowing both methods to share the same online 3DGS reconstruction pipeline.  
ActiveGS retains its original planner, which does not constrain the pitch orientation of the viewpoints, resulting in trajectories that are infeasible for the quadrotor.

\begin{table}[t]
\centering
\caption{
Parameters of the Proposed System
}
\label{tab_params}
\begin{tabular}{l|ll}
\toprule
    &\textbf{Parameter}            					& \textbf{Value}		\\ 
\midrule
    \multicolumn{1}{l|}{\multirow{6}{*}{\rotatebox{90}{Online 2DGS}}}    
                            & Initialization rejection distance ($\gamma$)             & \SI{5}{\centi\meter}     \\
    \multicolumn{1}{c|}{}   & Training iterations for a new frame ($K_{\mathrm{new}}$)             & $60$ \\
    \multicolumn{1}{c|}{}   & Training iterations for replay ($K_{\mathrm{replay}}$)         & $40$ \\
    \multicolumn{1}{c|}{}   & Depth rendering loss weight ($\lambda_\mathrm{t,D}$)      & $5.0$ \\
    \multicolumn{1}{c|}{}   & Depth distortion loss weight ($\lambda_\mathrm{t,D_d}$)   & $1.0$ \\
    \multicolumn{1}{c|}{}   & Normal consistency loss weight ($\lambda_\mathrm{t,N_c}$) & $0.05$ \\
\midrule
    \multicolumn{1}{l|}{\multirow{5}{*}{\rotatebox{90}{Reliability}}}    
                            & Near-edge-on observation threshold ($w_{\min}$)               & $0.1$ \\
    \multicolumn{1}{c|}{}   & Reconstruction loss weight ($\lambda_L$)              & $1.0$ \\
    \multicolumn{1}{c|}{}   & Cumulative transmittance weight ($\lambda_T$)         & $1.2$ \\
    \multicolumn{1}{c|}{}   & Depth rendering loss weight ($\lambda_\mathrm{r,D}$)     & $3.0$ \\
    \multicolumn{1}{c|}{}   & Normal rendering loss weight ($\lambda_\mathrm{r,N_r}$)  & $0.15$ \\
\bottomrule 
\end{tabular}

\end{table}

All methods are evaluated on eight scenes from Replica~\cite{replica19arxiv}.
Each experiment selects 200 frames, and the resulting reconstruction models are collected for further evaluation. 
The parameter values used in our system are summarized in Table~\ref{tab_params}.



\subsection{Evaluation Metrics}

We evaluate reconstruction quality from both visual and geometric perspectives. 
Visual quality is assessed using PSNR, SSIM, and LPIPS, which measure image fidelity, structural similarity, and perceptual discrepancy, respectively. 
Geometric quality is evaluated using the Depth L1 (D-L1) [\SI{}{\centi\meter}], Accuracy (Acc)[\SI{}{\centi\meter}], Chamfer Distance (CD)[\SI{}{\centi\meter}], and F-score. 
Specifically, D-L1 measures depth rendering accuracy.
Acc and CD are computed from nearest-neighbor distances between the reconstructed and ground-truth meshes.
F-score is the harmonic mean of precision and recall under a \SI{5}{\centi\meter} distance threshold.

For clarity, we also refer to PSNR, SSIM, LPIPS, and D-L1 as rendering quality metrics. 
In contrast, Acc, CD, and F-score are referred to as mesh quality metrics. 
All rendering metrics are evaluated on a fixed set of ground-truth viewpoints shared by all methods. 
To assess mesh quality, 
we extract meshes from the Gaussian models by fusing rendered depth maps through TSDF integration, without additional geometry optimization.


\subsection{Active Scene Reconstruction}
\label{subsec_active_reconstruction}

\begin{table}[t]
\centering
\caption{
Overall Evaluation Results
}
\label{tab_overall_result}
\begin{threeparttable}
    \setlength{\tabcolsep}{4.0pt}
    \scriptsize
\begin{tabular}{@{}c|ccccccc@{}}
\toprule
    Metrics 
    & \multicolumn{3}{c|}{Visual Quality} & \multicolumn{4}{c}{Geometric Quality} 
\\ 
\midrule
    Methods
    & PSNR$\uparrow$ & SSIM$\uparrow$ & \multicolumn{1}{c|}{LPIPS$\downarrow$} 
    & D-L1$\downarrow$ 
    & Acc$\downarrow$ & CD$\downarrow$
    & \multicolumn{1}{c}{F-score$\uparrow$ }
\\ 
\midrule
    Ours
    & \textbf{36.87} & \textbf{0.979} & \multicolumn{1}{c|}{ \textbf{0.036} } 
    & \textbf{0.25} 
    & \textbf{1.35} &  1.67 & \textbf{0.970} 
\\
    ActiveGS~\cite{jin2025ral}
    & 32.81 & 0.956 & \multicolumn{1}{c|}{ 0.066 } 
    & 0.70 
    & 1.47 &  \textbf{1.34} & 0.965 
\\
    GauSS-MI~\cite{YuhanRSS25}
    & 34.70 & 0.966 & \multicolumn{1}{c|}{ 0.071 } 
    & 1.47 
    & 2.28 &  2.41 & 0.928 
\\
    FisherRF~\cite{jiang2025fisherrf}
    & 34.57 & 0.966 & \multicolumn{1}{c|}{ 0.070 } 
    & 1.44 
    & 2.27 &  2.39 & 0.929 
\\
\bottomrule
\end{tabular}

\begin{tablenotes}
    \item Evaluation is conducted on all eight \textit{Office} and \textit{Room} scenes in the Replica dataset. Each method is evaluated three times on each scene, and the results are averaged over the three runs and all scenes. 
    Detailed per-scene results are reported in Table~\ref{tab_result_per_scene}.
\end{tablenotes}

\end{threeparttable}
\end{table}

\begin{table}[t]
\centering
\caption{
Per-Scene Evaluation Results on Replica
}
\label{tab_result_per_scene}
\begin{threeparttable}
    \setlength{\tabcolsep}{2.4pt}
    \scriptsize
\begin{tabular}{@{}c|c|cccccccc@{}}
\toprule
    Methods $^1$
    & Metrics $^2$
    & Of0 & Of1 & Of2 & Of3 & Of4 & R0 & R1 & R2 $^3$
    \\ 
\midrule
    \multicolumn{1}{c|}{\multirow{3}{*}{Ours}}
    & 
    PSNR$\uparrow$
    & \textbf{37.45} & \textbf{34.35} & \textbf{35.38}
    & \textbf{35.98} & \textbf{38.67} &  \textbf{38.21}
    & 37.91 & \textbf{36.99}
    \\
    \multicolumn{1}{c|}{}
    &
    D-L1$\downarrow$
    & \textbf{0.21} & \textbf{0.13} & \textbf{0.24}
    & \textbf{0.32} & \textbf{0.24} &  \textbf{0.29}
    & \textbf{0.23} & \textbf{0.31}
    \\
    \multicolumn{1}{c|}{}
    &
    F-score$\uparrow$
    & \textbf{0.987} & \textbf{0.969} & \textbf{0.962}
    & \textbf{0.952} & \textbf{0.964} &  0.964
    & \textbf{0.980} & \textbf{0.983}
    \\
\midrule
    \multicolumn{1}{c|}{\multirow{3}{*}{ActiveGS~\cite{jin2025ral}}}
    & 
    PSNR$\uparrow$
    & 33.87 & 30.03 & 30.52
    & 32.58 & 34.49 &  34.24
    & 34.85 & 31.91
    \\
    \multicolumn{1}{c|}{}
    &
    D-L1$\downarrow$
    & 0.52 & 0.50 & 0.70
    & 0.88 & 0.74 &  0.65
    & 0.61 & 1.03
    \\
    \multicolumn{1}{c|}{}
    &
    F-score$\uparrow$
    & 0.981 & 0.965 & 0.952 & \textbf{0.952} & 0.943 
    &  \textbf{0.978} & 0.978 & 0.969
    \\
\midrule
    \multicolumn{1}{c|}{\multirow{3}{*}{GauSS-MI~\cite{YuhanRSS25}}}
    & 
    PSNR$\uparrow$
    & 35.17 & 33.42 & 31.27
    & 31.26 & 37.14 &  36.12
    & 37.55 & 35.65
    \\
    \multicolumn{1}{c|}{}
    &
    D-L1$\downarrow$
    & 1.30 & 1.27 & 1.67
    & 1.67 & 1.50 &  1.54
    & 1.26 & 1.54
    \\
    \multicolumn{1}{c|}{}
    &
    F-score$\uparrow$
    & 0.961 & 0.935 & 0.893
    & 0.901 & 0.921 &  0.926
    & 0.961 & 0.925
    \\
\midrule
    \multicolumn{1}{c|}{\multirow{3}{*}{FisherRF~\cite{jiang2025fisherrf}}}
    & 
    PSNR$\uparrow$
    & 35.04 & 33.40 & 30.21
    & 30.96 & 36.83 & 36.48
    & \textbf{38.23} & 35.43
    \\
    \multicolumn{1}{c|}{}
    &
    D-L1$\downarrow$
    & 1.33 & 1.28 & 1.43
    & 1.69 & 1.44 &  1.50
    & 1.15 & 1.70
    \\
    \multicolumn{1}{c|}{}
    &
    F-score$\uparrow$
    & 0.959 & 0.937 & 0.899
    & 0.905 & 0.926 &  0.926
    & 0.967 & 0.912
    \\
\bottomrule
\end{tabular}

\begin{tablenotes}
    \item[1] Each method is evaluated three times on each scene, and the results are averaged.
    \item[2] The metrics cover color rendering, depth rendering, and mesh quality.
    \item[3] `Of' denotes \textit{Office} scenes, and `R' denotes \textit{Room} scenes in the Replica dataset.
\end{tablenotes}
\end{threeparttable}

\vspace{-0.3cm}
\end{table}

\begin{figure*}[t] 
    \centering
    \includegraphics[width=\linewidth]{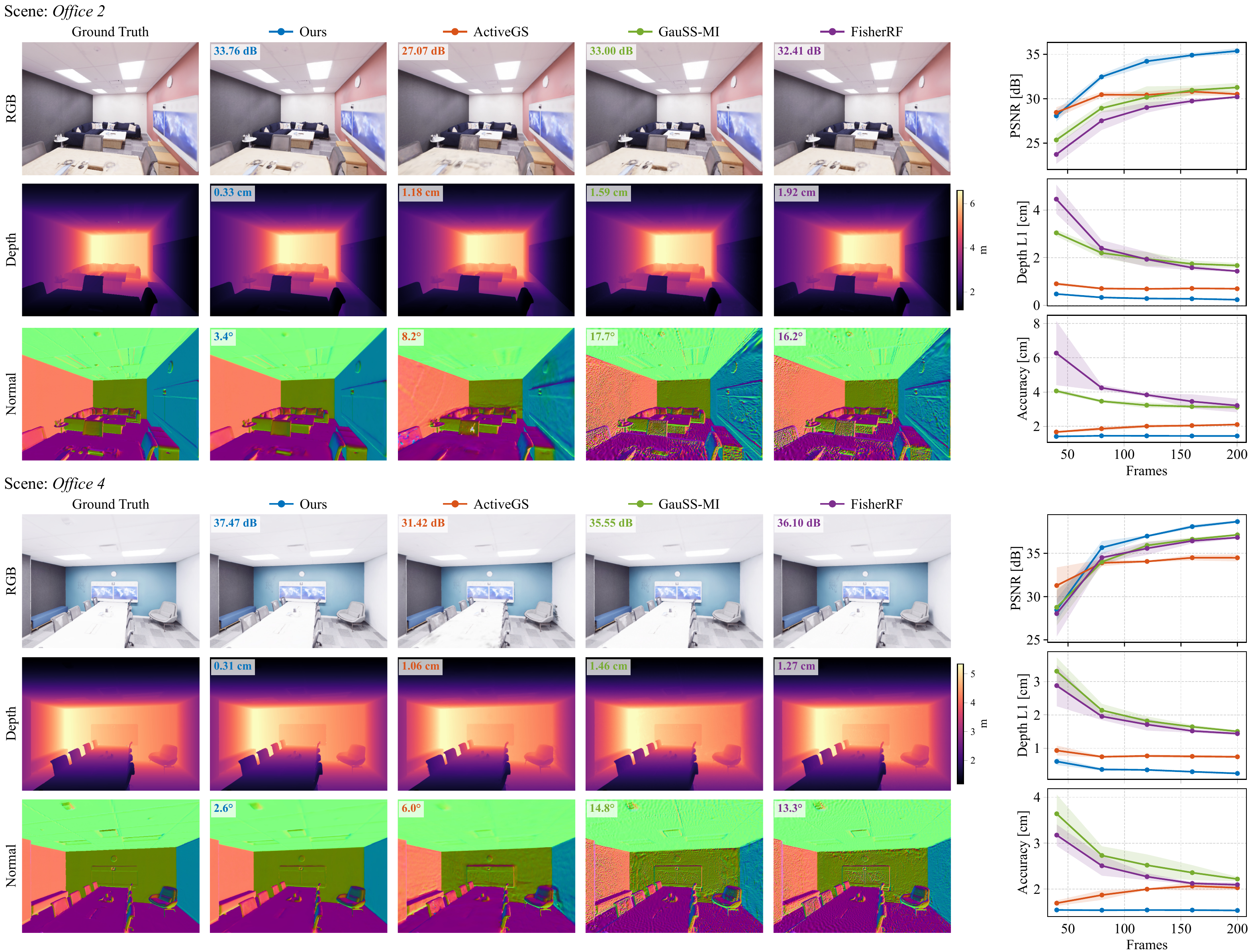}
    \caption{
        Novel view synthesis results compared to the ground truth, together with reconstruction quality over frames, for the scenes \textit{Office~2} and \textit{Office~4}. 
        For each scene, the rendered color, depth, and normal images are presented, and each rendering is annotated with its PSNR~[dB], Depth L1~[cm], and mean normal angular error~[$^\circ$], respectively.
        The curves on the right report PSNR, Depth L1, and reconstruction accuracy as the number of collected frames increases, averaged over three runs, with the shaded areas indicating the variation across runs.
    }
    \label{fig_novel_render}
\end{figure*}




The comparison results averaged over eight scenes are summarized in Table~\ref{tab_overall_result},
and Table~\ref{tab_result_per_scene} details the per-scene performance on color rendering, depth rendering, and mesh quality. 
Overall, our method achieves the best performance in most evaluations across the two tables, with only three exceptions: ActiveGS achieves a lower overall CD and a higher F-score in the scene \textit{Room~0}, and FisherRF achieves a higher PSNR in \textit{Room~1}.
Additionally, ActiveGS ties with our method for the best F-score in \textit{Office~3}.
Although our method is not consistently the best on every individual metric, it achieves the strongest overall performance across the evaluated scenes. 
We attribute this to the reliability model introduced in Section~\ref{subsec_reliability}, where the reconstruction loss $L_k$ jointly measures color, depth, and normal residuals.
This enables the system to identify viewpoints with insufficient reconstruction quality and prioritize observations that provide greater information gain. 
Meanwhile, the online reconstruction jointly optimizes photometric and depth losses with geometric regularizers, allowing the acquired observations to contribute to both appearance and geometry throughout the reconstruction process.

The comparison also highlights distinct behaviors among the baselines. 
ActiveGS achieves the best CD, indicating favorable overall surface reconstruction despite its lower rendering quality. 
However, its higher D-L1 and Acc and relatively lower F-score reveal less accurate depth reconstruction and surface alignment.
GauSS-MI and FisherRF, in contrast, achieve stronger visual results than ActiveGS but exhibit larger depth and mesh errors. 
Our method achieves a better balance between visual and geometric quality by jointly considering both during view selection, rather than relying solely on scene coverage or visual information gain.

The qualitative comparison on the left-hand side of Figure~\ref{fig_novel_render} further supports these observations. 
ActiveGS provides relatively complete geometric coverage but produces less accurate renderings, whereas GauSS-MI and FisherRF preserve better visual fidelity at the expense of geometric accuracy. 
Our method achieves a more consistent reconstruction, with sharper textures and cleaner object boundaries.

The right-hand side of Figure~\ref{fig_novel_render} shows the reconstruction quality as the active reconstruction progresses, with the three plots corresponding to the color, depth, and mesh quality shown in the three rows on the left, respectively.
ActiveGS improves rapidly during the early stage but begins to saturate after approximately $80$ frames, with its accuracy slightly degrading thereafter. 
This behavior in the initial stage can be largely attributed to its original view planner, which neither accounts for quadrotor trajectory feasibility nor constrains the camera pitch, allowing it to select downward-facing viewpoints that are not reachable by the quadrotor. 
Such viewpoints provide broad observations of horizontal surfaces and can therefore yield strong initial coverage, but this advantage does not persist as the reconstruction progresses. 
In contrast, our method evaluates candidate viewpoints only after trajectory planning and feasibility checking.
Although this constraint may limit the initial reconstruction speed, the proposed reliability-aware view selection continues to identify informative observations and steadily improves the reconstruction as more frames are acquired.



\subsection{Efficiency Study}

\begin{figure}[t] 
    \centering
    \includegraphics[width=0.94\linewidth]{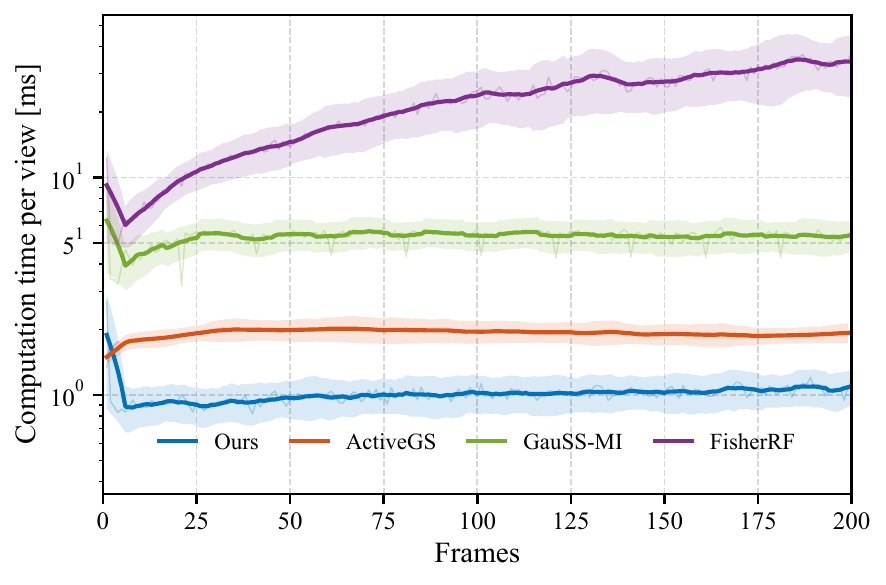}
    \vspace{-0.3cm}
    \caption{
        Candidate evaluation time over the active reconstruction process, averaged over three runs across eight scenes. 
    }
    \label{fig_runtime}
    \vspace{-0.4cm}
\end{figure}

\begin{table}[t]

\centering
\caption{
    Computational Efficiency and Model Size
}
\label{tab_efficiency}
\begin{threeparttable}
    \scriptsize
\begin{tabular}{@{}c|ccc@{}}
\toprule
    Methods
    & Time [ms]$^1\downarrow$ & fps [s$^{-1}$]$\uparrow$ &  Model Size [MB]$\downarrow$
\\
\midrule
    Ours
    & \textbf{1.01} & \textbf{990.6} & 10.4
\\
    ActiveGS~\cite{jin2025ral}
    & 1.93 & 518.8 & 24.7
\\
    GauSS-MI~\cite{YuhanRSS25}
    & 5.34 & 187.2 & \textbf{9.0}
\\
    FisherRF~\cite{jiang2025fisherrf}
    & 22.03 & 45.4 & 9.4
\\
\bottomrule
\end{tabular}
\begin{tablenotes}
    \item[1]
    Computation time for a single candidate viewpoint, averaged over all candidate evaluations from 24 complete active reconstruction runs, with 200 frames per run.
\end{tablenotes}
\end{threeparttable}

\vspace{-0.7cm}
\end{table}


This subsection evaluates the efficiency of the proposed method in terms of computational cost and model size.

We record the computation time of all methods throughout the complete active reconstruction process and summarize the results in Figure~\ref{fig_runtime}. 
As the reconstruction progresses, the computation time of our method remains relatively stable, whereas the baselines exhibit higher overhead. 
In particular, FisherRF shows an approximately linear increase in computation time with the number of frames, 
as it requires rendering over all observed frames at each view selection step,
making it unsuitable for real-time active reconstruction.
The higher computation time observed during the initial frames for our method, GauSS-MI, and FisherRF is likely due to the shared simulation and trajectory-planning modules, which introduce additional initialization overhead. 
Once initialized, their computation time remains relatively stable throughout the reconstruction process.

To provide a quantitative comparison, Table~\ref{tab_efficiency} reports the average computation time for candidate evaluation, the corresponding rate in frames per second (fps), and the resulting model size. 
Our method achieves the highest rate of \SI{990.6}{\fps}, substantially outperforming all three baselines. 
This low computational cost is particularly beneficial for active reconstruction, 
where view selection and map updates must be performed online throughout the reconstruction process.
In terms of model size, our method maintains a compact $10.4$ MB map, substantially smaller than ActiveGS and comparable to GauSS-MI and FisherRF. 
The similar model sizes of GauSS-MI and FisherRF are expected, as they share the same online reconstruction pipeline, with the remaining difference arising from their different selections of viewpoints. 

Overall, our method achieves superior reconstruction quality with the lowest computational cost and a compact model size, 
as summarized in Figure~\ref{fig_sactter_methods}.



\section{Conclusions}

This paper presents an efficient active reconstruction system that balances visual and geometric reconstruction quality, computational efficiency, and model compactness.
We develop an efficient incremental 2DGS mapping pipeline for real-time high-quality reconstruction, and introduce a probabilistic reliability model that characterizes the view-dependent reconstruction quality of individual 2D Gaussian splats based on their explicit surface orientation. 
Building on this model, we formulate 2D GauSS-MI to estimate the information provided by novel views and guide view selection according to both visual and geometric quality. 
Extensive experiments on eight Replica scenes demonstrate that the proposed framework achieves favorable overall reconstruction quality with low computational cost and a compact model size. 






\section*{Acknowledgment}
The authors gratefully acknowledge Yixi Cai for insightful and valuable suggestions. 


\bibliographystyle{IEEEtran}
\bibliography{root}


\end{document}